\pdfoutput=1
\newcommand{\Forward}{Forward Augmentation\xspace}
\newcommand{\Backward}{Backward Augmentation\xspace}
\newcommand{\Roundtrip}{Round-trip Augmentation\xspace}

\newcommand{\Sreal}{$\mathbf{S}$\xspace}
\newcommand{\Treal}{$\mathbf{T}$\xspace}
\newcommand{\Spseudo}{$\mathbf{S}_\text{pseudo}$\xspace}
\newcommand{\Tpseudo}{$\mathbf{T}_\text{pseudo}$\xspace}
\newcommand{\Ppseudo}{$\mathbf{P}_\text{pseudo}$\xspace}

\newcommand{\Original}{Original\xspace}
\newcommand{\Rewriter}{Rewriter\xspace}
\newcommand{\Gold}{Gold\xspace}

\newcommand{\citeg}[1]{\citeauthor{#1}'s \citeyearpar{#1}} % cite genitive: Martinez et al's (2012)...

\documentclass[11pt]{article}

\usepackage[]{ACL2023}

\usepackage{times}
\usepackage{latexsym}
\usepackage{booktabs}
\usepackage{graphicx}
\usepackage{tabularx}
\usepackage{amsmath}
\usepackage{algorithm}
\usepackage{algpseudocode}
\usepackage{xspace}
\usepackage{caption}
\usepackage{subcaption}

\definecolor{highlight}{HTML}{ff8000}

\usepackage[T1]{fontenc}
\usepackage[utf8]{inputenc}

\usepackage{microtype}

\usepackage{inconsolata}

\title{Exploiting Biased Models to De-bias Text: A Gender-Fair Rewriting Model}

\author{Chantal Amrhein$^{1}$\thanks{{ } Work done during an internship at Textshuttle.} \hspace{0.5cm}  Florian Schottmann$^{2,3}$ \hspace{0.5cm}  Rico Sennrich$^{1,4}$ \hspace{0.5cm} Samuel Läubli$^{1,2}$  \medskip \\
  $^1$University of Zurich, 
  $^2$Textshuttle,  
  $^3$ETH Zurich,  
  $^4$University of Edinburgh \medskip \\ 
  \texttt{\{amrhein,sennrich\}@cl.uzh.ch, \{schottmann,laeubli\}@textshuttle.ai}}

\begin{document}
\maketitle
\begin{abstract}
Natural language generation models reproduce and often amplify the biases present in their training data. Previous research explored using sequence-to-sequence rewriting models to transform biased model outputs (or original texts) into more gender-fair language by creating pseudo training data through linguistic rules. However, this approach is not practical for languages with more complex morphology than English. We hypothesise that creating training data in the reverse direction, i.e. starting from gender-fair text, is easier for morphologically complex languages and show that it matches the performance of state-of-the-art rewriting models for English. To eliminate the rule-based nature of data creation, we instead propose using machine translation models to create gender-biased text from real gender-fair text via round-trip translation. Our approach allows us to train a rewriting model for German without the need for elaborate handcrafted rules. The outputs of this model increased gender-fairness as shown in a human evaluation study.\footnote{We publicly release our data and code
here: \url{https://github.com/textshuttle/exploiting-bias-to-debias}}
\end{abstract}

\section{Introduction}
\label{sec:intro}

\begin{figure}[t!]
    \centering
    \begin{subfigure}[b]{\columnwidth}
        \centering
        \includegraphics[width=8cm]{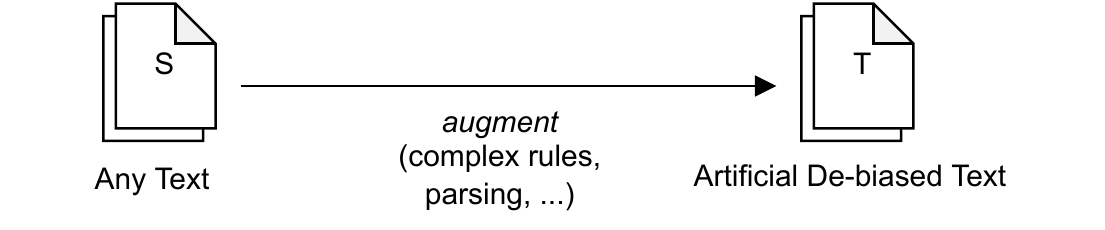}
        \captionsetup{justification=centering}
        \caption{\Forward\\\citep{vanmassenhove-etal-2021-neutral, sun-etal-2021-they}}
        \label{fig:approach_forward}
    \end{subfigure}
    \hfill
    \vspace{-1mm}
    \begin{subfigure}[b]{\columnwidth}
        \centering
        \includegraphics[width=7cm, trim=15 0 0 0,clip]{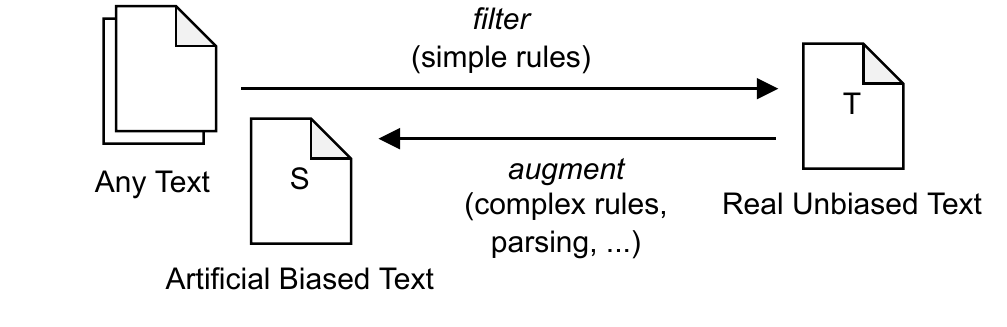}
        \captionsetup{justification=centering}
        \caption{\Backward (this work)}
        \label{fig:approach_backward}
    \end{subfigure}
    \hfill
    \vspace{-2mm}
    \begin{subfigure}[b]{\columnwidth}
        \centering
        \includegraphics[width=7cm, trim=15 0 0 0,clip]{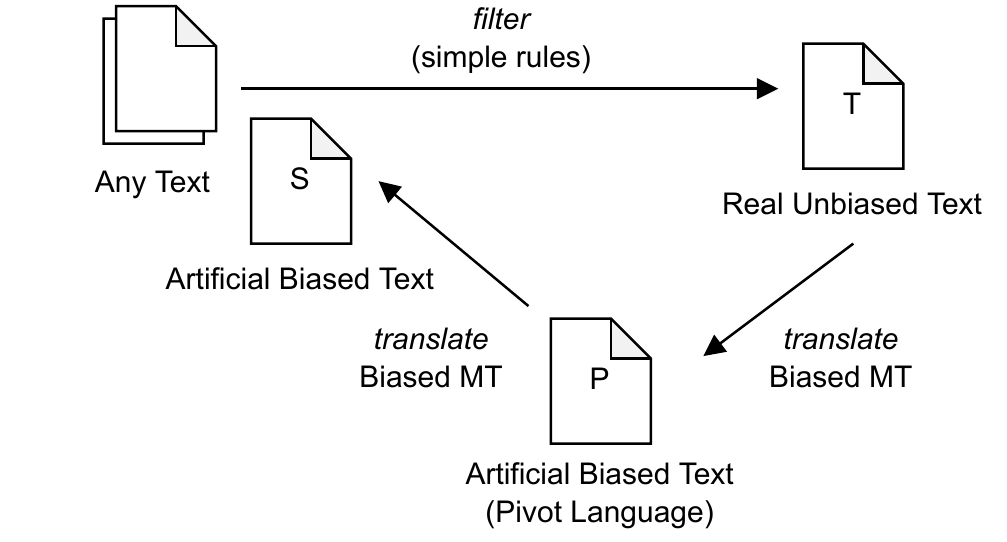}
        \captionsetup{justification=centering}
        \caption{\Roundtrip (this work)}
        \label{fig:approach_roundtrip}
    \end{subfigure}
    \caption{De-biasing rewriters can be implemented as neural sequence-to-sequence models trained on source (\Sreal) to target (\Treal) text examples. Previous work creates artificial \Treal from real \Sreal through complex augmentation (\subref{fig:approach_forward}). We propose to use real \Treal and generate artificial \Sreal to accommodate morphologically complex languages and avoid target-side noise (\subref{fig:approach_backward}). Furthermore, we show that by leveraging biased off-the-shelf machine translation (MT) models, complex rules can be avoided altogether to generate training data for de-biasing rewriters (\subref{fig:approach_roundtrip}).}
    \label{fig:approach}
\end{figure}

From facial recognition to job matching and medical diagnosis systems, numerous real-world applications suffer from machine learning models that are discriminative towards minority groups based on characteristics such as race, gender, or sexual orientation \citep{mehrabi-2021-ai-fairness}. In natural language processing (NLP), gender bias is a particularly significant issue \citep{sheng-etal-2021-societal}. While research in psychology, linguistics and education studies demonstrates that inclusive language can increase the visibility of women \citep{horvath-etal-2016-does,tibblin-etal-2022-there} and encourage young individuals of all genders to pursue stereotypically gendered occupations \citep{vervecken-etal-2013-changing,vervecken-etal-2015-warm} without sacrificing comprehensibility \citep{friedrich-heise-2019-does}, state-of-the-art text generation models still overproduce masculine forms and perpetuate gender stereotypes \citep{stanovsky-etal-2019-evaluating, nadeem-etal-2021-stereoset, renduchintala-williams-2022-investigating}.

There is a variety of work on correcting gender bias in generative models. Some approaches include curating balanced training data \citep{saunders-etal-2020-neural}, de-biasing models with modifications to the training algorithms \citep{choubey-etal-2021-gfst}, and developing better inference procedures \citep{saunders-etal-2022-first}. The focus of our work lies on so-called rewriting models, yet another line of de-biasing research that revolves around models that map any input text (e.g., the output of a biased generative model) to a gender-fair version of the same text. The main challenge here is training data: due to the lack of large amounts of parallel biased-unbiased text segments, previous work (Section~\ref{sec:background}) produces the latter through handcrafted rules (\Forward, Figure~\ref{fig:approach_forward}).

\begin{figure}
    \centering
    \includegraphics[width=0.5\textwidth]{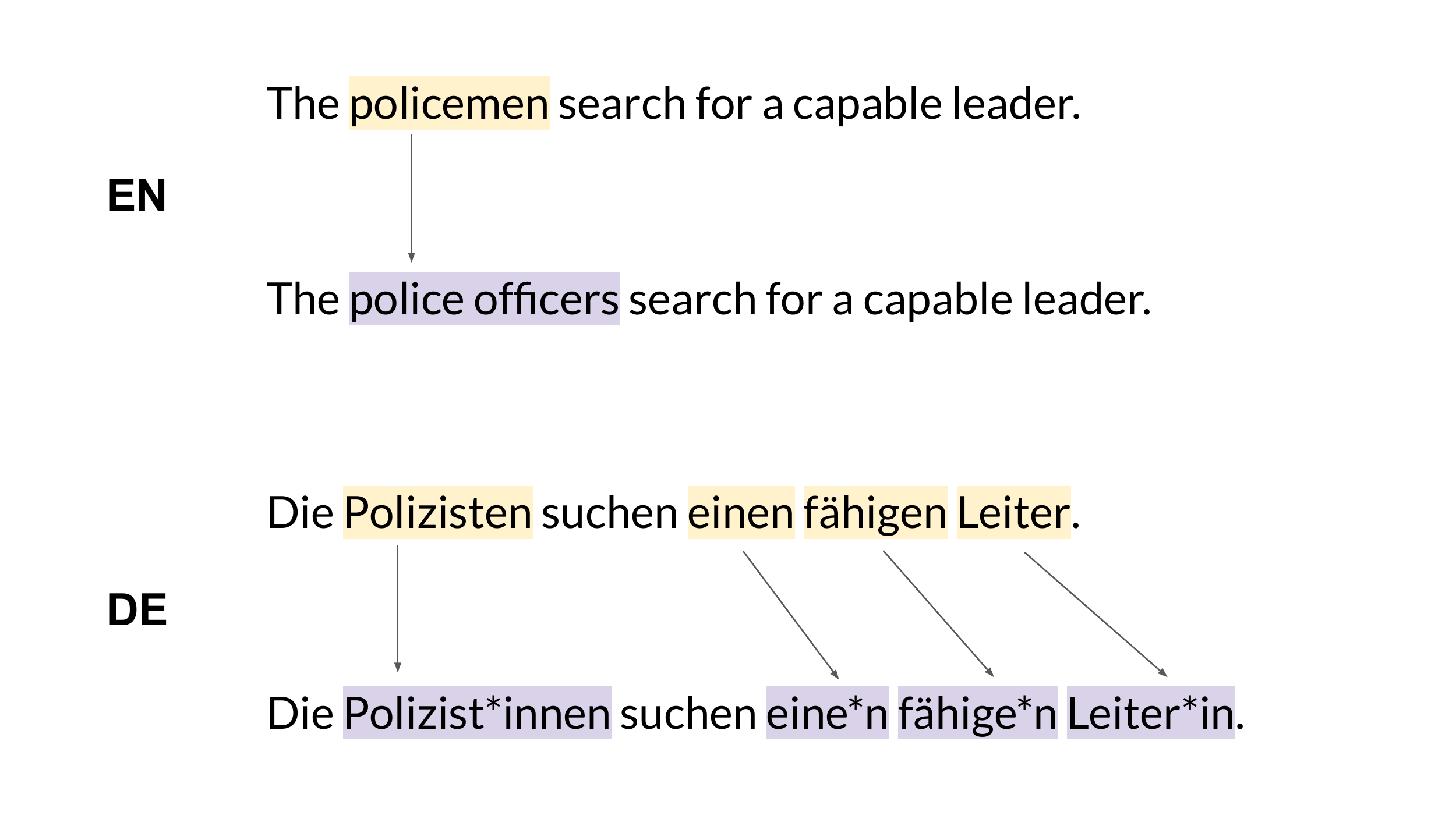}
    \caption{Gendered words (yellow) that need to be altered for gender-fairness (purple) in English (EN) and German (DE).
    }
    \label{fig:example}
\end{figure}

We identify two key problems with the \Forward paradigm. First, the rule-based de-biasing of real-world text comes at a risk of introducing target-side noise, which tends to degrade output quality more than source-side noise \citep{khayrallah-koehn-2018-impact,bogoychev-sennrich-2019-backtranslation}. Second, while already intricate for English, it is likely even harder to define de-biasing rules for more morphologically complex languages with grammatical gender (Figure~\ref{fig:example}). An approach proposed by \citet{diesner-etal-2022-supporting}, for example, requires morphological, dependency and co-reference analysis, as well as named entity recognition and a word inflexion database.

In this paper, we propose two modifications to the prevalent data augmentation paradigm and we find that biased models can be used to train de-biasing rewriters in a simple yet effective way. Our three main contributions are:

\begin{itemize}
    \item We reverse the data augmentation direction (\Backward, Figure~\ref{fig:approach_backward}). By using human-written unbiased segments filtered from large monolingual corpora as target-side data, we train a neural rewriting model that matches or outperforms the word error rate (WER) of two strong \Forward baselines in English (Section~\ref{sec:backward_eval_auto}).
    \item We dispose of handcrafted de-biasing rules (\Roundtrip, Figure~\ref{fig:approach_roundtrip}). By leveraging biased off-the-shelf NLP models, we train a neural rewriting model that outperforms the WER of a heavily engineered rule-based system in German (Section~\ref{sec:roundtrip_eval_auto}).
    \item We test our best model with potential stakeholders. In our human evaluation campaign, participants rated the outputs of our German rewriter model as more gender-fair than the original (biased) input texts (Section~\ref{sec:human}).
\end{itemize}

\section{Background}
\label{sec:background}

Gender-fair\footnote{We define ``gender-fair'' as marking all genders explicitly; in contrast to ``gender-neutral'' where gender is deliberately \textit{not} marked. For ease of reading, we only use the term ``gender-fair'' in this paper, even though our English models produce gender-neutral text.} rewriting is a conditional text generation problem. It can be approached with conventional sequence-to-sequence models trained on large amounts of parallel data, i.e., biased segments \Sreal and their gender-fair counterparts \Treal. Since such corpora do not exist in practice,\footnote{One exception: \citet{qian-etal-2022-perturbation} recently published a 100k segment crowd-sourced parallel corpus for English.} \citet{sun-etal-2021-they} and \citet{vanmassenhove-etal-2021-neutral} create artificial gender-fair target segments \Tpseudo from existing source segments \Sreal with a rule-based de-biasing pipeline (\Forward, Figure~\ref{fig:approach_forward}). The sequence-to-sequence model trained on the outputs of this pipeline removes the need for computationally expensive toolchains (such as dependency parsing) at runtime and increases robustness towards noisy inputs \citep{vanmassenhove-etal-2021-neutral}.

\subsection{Rule-based De-biasing for English}
\label{sec:background_english}

Converting \Sreal to \Tpseudo is relatively straightforward for languages like English, which only expresses social gender in pronouns and a small set of occupation nouns. A simple dictionary lookup is often sufficient to produce a gender-fair variant of the biased text, except for two issues:

\begin{itemize}
    \item[] \small \textbf{Issue 1: Rewritings that affect other dependencies}, e.g.\ when rewriting third-person subject pronouns in English, verbs in present tense need to be pluralised (e.g.\ ``she knows'' rewritten as ``they know'').
    \item[] \small \textbf{Issue 2: Ambiguities that need to be resolved from context}, e.g.\ English ``her'' can be a possessive pronoun (rewritten as ``their'') or a personal pronoun (rewritten as ``them'').
\end{itemize}

\noindent These issues are tractable for English because they only happen in a limited number of cases that can be covered with a limited number of rules. \citet{sun-etal-2021-they} solve \textit{Issue 1} based on part-of-speech, morphological and dependency information, and \textit{Issue 2} by scoring different variants with a language model. \citet{vanmassenhove-etal-2021-neutral} solve \textit{Issue 1} using a grammatical error correction tool and \textit{Issue 2} using part-of-speech, morphological and dependency information. Both \citet{sun-etal-2021-they} and \citet{vanmassenhove-etal-2021-neutral} train Transformer models \citep{vaswani-etal-2017} using the original texts as the source and the de-biased augmentations as target data (\Forward), and achieve WER below 1\% on several test sets.

\subsection{Rule-based De-biasing for Other Languages}
\label{sec:background_other}

In languages with more complex morphology, \textit{Issue 1} is much more prevalent than in English but \textit{Issue 2} is even more challenging because it requires animacy prediction: A direct application of \citeg{vanmassenhove-etal-2021-neutral} ``first rule-based, then neural'' approach to Spanish \citep{jain-etal-2021-generating} results in a model that does not distinguish between human referents and objects. Similarly, \citet{alhafni-etal-2022-user}\footnote{The latest continuation of a series of work for gender-fair rewriting in Arabic \citep{habash-etal-2019-automatic, alhafni-etal-2020-gender}.} train an end-to-end rewriting system for Arabic but their pipeline for creating training data requires labelled data to train an additional classification model to identify gendered words. Both of these works only focus on a binary interpretation of gender. Non-sequence-to-sequence approaches have also been explored \citep{zmigrod-etal-2019-counterfactual, diesner-etal-2022-supporting} but required extensive linguistic tools such as morphological, dependency and co-reference analysis, named entity recognition and word inflexion databases.

\subsection{Round-trip Translation}
\label{sec:background_roundtrip}

Previous work employed round-trip translations to create pseudo data for automatic post-editing \citep{junczys-dowmunt-grundkiewicz-2016-log, freitag-etal-2019-ape, voita-etal-2019-context}, grammatical error correction \citep{madnani-etal-2012-exploring, lichtarge-etal-2019-corpora} or paraphrasing \citep{mallinson-etal-2017-paraphrasing, iyyer-etal-2018-adversarial,fabbri-etal-2021-improving, cideron-et-al-2022-vec2text}. While such uses of round-trip translations exploit the fact that machine translations can be diverse and can contain accuracy and fluency errors, we are the first to exploit them for their social biases.

\section{\Backward}
\label{sec:backward}

\begin{table*}
    \centering
    \small
    \begin{tabular}{lccccc}
    & \multicolumn{2}{c}{\textbf{\citeauthor{sun-etal-2021-they}}} & \multicolumn{3}{c}{\textbf{\citeauthor{vanmassenhove-etal-2021-neutral}}}  \\  \cmidrule(lr){2-3} \cmidrule(lr){4-6} 
         &  \textbf{non-gendered} & \textbf{gendered} & \textbf{OpenSubtitles} & \textbf{Reddit} & \textbf{WinoBias+} \\ \cmidrule(lr){2-2}  \cmidrule(lr){3-3} \cmidrule(lr){4-4} \cmidrule(lr){5-5} \cmidrule(lr){6-6}
    Source (no rewriting) & \textbf{0.00} & 10.72 & 14.03 & 10.85 & 8.70\vspace{2mm}\\
    \Forward:\\(a) \citet{vanmassenhove-etal-2021-neutral} & - & - & \phantom{1}0.43 & \phantom{1}0.75 & 0.09 \\
    (b) \citet{sun-etal-2021-they} &  \textbf{0.00} & \phantom{1}\textbf{0.57} & - & - & - \\
    (c) Reimplementation (a + b) &  \textbf{0.00} & \phantom{1}\textbf{0.42} & \phantom{1}\textbf{0.30} & \phantom{1}\textbf{0.46} & \textbf{0.05}\vspace{2mm}\\
    \Backward (this work) &  \textbf{0.00} & \phantom{1}\textbf{0.43} &  \phantom{1}\textbf{0.24} & \phantom{1}\textbf{0.40} & \textbf{0.04}
    \end{tabular}
    \caption{Tokenised WER (lower is better) of different rewriting approaches for English. Best systems (no other statistically significantly better) marked in bold; \Backward matches \Forward.}
    \label{tab:wer_english}
\end{table*}

We hypothesise that the data augmentation direction for gender-fair rewriters can be reversed (Figure~\ref{fig:approach_backward}) without a negative impact on quality. Our motivation is rooted in work on data augmentation for MT \citep{sennrich-2016-backtranslation}, where back-translation of monolingual target text tends to result in better quality than forward-translation of original source text \citep{khayrallah-koehn-2018-impact,bogoychev-sennrich-2019-backtranslation}. We use \Backward to train a gender-fair rewriter model for English and compare its performance to the \Forward approach proposed by \citet{vanmassenhove-etal-2021-neutral} and \citet{sun-etal-2021-they}.

\subsection{Method}
\label{sec:backward_method}

We propose to filter large monolingual corpora for gender-fair text \Treal and use a rule-based pipeline to derive artificially biased source text \Spseudo from \Treal. Based on this data, we can train a sequence-to-sequence model which maximises $p(\mathbf{T} | \mathbf{S}_\text{pseudo}, \boldsymbol{\theta})$, rather than $p(\mathbf{T}_\text{pseudo} | \mathbf{S}, \boldsymbol{\theta})$ as in previous work (Section~\ref{sec:background}).

\subsection{Experimental Setup}
\label{sec:backward_experiment}

\paragraph{Data} We extract English training data from OSCAR \citep{abadji-etal-2022-towards}, a large multilingual web corpus. For \Forward, we select segments that contain at least one biased word as \Sreal, following \citeg{vanmassenhove-etal-2021-neutral} and \citeg{sun-etal-2021-they} lookup tables  (Appendix~\ref{app:english_lookup}). For \Backward, we filter for segments that contain at least one of the corresponding gender-fair words in the lookup tables as \Treal. We filter out duplicates and noisy segments with \texttt{OpusFilter} \citep{aulamo-etal-2020-opusfilter} and then randomly subselect 5M segments each. For both models and as in previous work, we extend the training data by creating complementary source versions with only masculine forms, only feminine forms and copies of gender-fair targets and by adding additional non-gendered segments where no rewriting is necessary (amounting to 30\% of the total data). A full overview of the training data can be found in Appendix \ref{app:data-overview}.

\paragraph{Rule-based Processing} To be able to compare directly to previous work, we first reproduce the rule-based \Forward approach proposed by \citet{sun-etal-2021-they} and \citet{vanmassenhove-etal-2021-neutral} to create \Tpseudo from \Sreal. We combine their lookup tables (Appendix~\ref{app:english_lookup}) and re-implement their rules based on part-of-speech, morphological and dependency information via \texttt{spaCy}\footnote{\url{https://spacy.io/}} \citep{spacy}, both for resolving ambiguities and producing the correct number for verbs. We decide to follow \citet{sun-etal-2021-they} and use ``themself'' and not ``themselves'' as a gender-fair form of ``herself'' and ``himself''. Taking this implementation as a basis, we derive a \Backward pipeline by reversing the lookup tables and rules to map from \Treal to \Spseudo.

\paragraph{Model Architecture} Following \citet{sun-etal-2021-they} and \citet{vanmassenhove-etal-2021-neutral}, we train 6-layer encoder, 6-layer decoder Transformers \citep{vaswani-etal-2017} with 4 attention heads, an embedding and hidden state dimension of 512 and a feed-forward dimension of 1024. For optimization, we use Adam \citep{kingma-etal-2015} with standard hyperparameters and a learning rate of $5e-4$. We follow the Transformer learning schedule in \citet{vaswani-etal-2017} with a linear warmup over 4,000 steps. The only differences to \citet{sun-etal-2021-they} and \citet{vanmassenhove-etal-2021-neutral} are that we train our models with \texttt{sockeye 3} \citep{hieber-etal-2022-sockeye3} and use a smaller joint byte-pair vocabulary \citep{sennrich-etal-2016-neural} of size 8k computed with \texttt{SentencePiece} \citep{kudo-richardson-2018-sentencepiece}.

\subsection{Automatic Evaluation}
\label{sec:backward_eval_auto}

\paragraph{Test Sets} We benchmark our models with the test sets published in conjunction with our baselines:

\begin{itemize}
    \item{\textbf{\citet{sun-etal-2021-they}}}: Two test sets (gendered/non-gendered) with 500 sentence pairs each, from five different domains: Twitter, Reddit, news articles, movie quotes and jokes. For the gendered version, there are balanced numbers of sentences with feminine and masculine pronouns for each domain. The non-gendered source texts do not contain any forms that need to be rewritten and should not be changed.
    \item{\textbf{\citet{vanmassenhove-etal-2021-neutral}}}: Three test sets from three different domains: OpenSubtitles \citep[500 sentence pairs]{lison-tiedemann-2016-OpenSubtitles2016}, Reddit  \citep[500 sentence pairs]{baumgartner-etal-2020-pushlift}, and WinoBias+ \citep[3,167 sentence pairs]{zhao-etal-2018-gender}. Each test set has a balanced amount of gender-fair pronoun types.
\end{itemize}

\noindent We manually double-check the target side of all test sets from previous work and if necessary correct sporadic human annotation errors. The test sets used by \citet{vanmassenhove-etal-2021-neutral} also cover grammatical error corrections outside the scope of gender-fair rewriting. To restrict evaluation to the phenomenon of interest, we produce a target side version that only covers gender-fair rewriting. Note that this means that the model outputs by \citet{vanmassenhove-etal-2021-neutral} will perform slightly worse on this version of the test set than reported in their paper because this model also makes such additional grammatical corrections. We revert tokenization and change ``themselves'' to ``themself'' in the model outputs of \citet{vanmassenhove-etal-2021-neutral} to be able to compare them against our models' outputs and our references.

\paragraph{Method} We evaluate our English model outputs and compare them to previous work with tokenised\footnote{\label{ftn:sacremoses}\url{https://github.com/alvations/sacremoses}} WER based on the Python package \texttt{jiwer}\footnote{\label{ftn:jiwer}\url{https://github.com/jitsi/jiwer}}. We compute statistical significance $p<0.05$ with
paired bootstrap resampling \citep{koehn-2004-statistical}, sampling 1,000 times
with replacement.

\paragraph{Results} Results are shown in Table \ref{tab:wer_english}. 
\Backward matches the low WER of the original as well as our combined reproduction of the \Forward models by \citet{sun-etal-2021-they} and \citet{vanmassenhove-etal-2021-neutral}, and performs slightly better than previous work on OpenSubtitles and Reddit and WinoBias+.

\section{\Roundtrip}
\label{sec:roundtrip}

Artificially biasing gender-fair target segments rather than de-biasing gender-biased source segments is especially useful for languages with grammatical gender and more complex morphology than English. Taking German as a running example, we would need some form of animacy prediction in \Forward to transform ambiguous nouns such as ``Leiter'' only if they refer to a person (``leader'') and not to an object (``ladder''). In \Backward, we do not need an animacy prediction model since this information is implicitly encoded in the gender-fair forms, as seen in Figure \ref{fig:example}. Nevertheless, defining rules for mapping gender-fair segments to gender-biased segments or vice versa requires expert knowledge and will likely never completely cover morphologically complex languages.

As an alternative to handcrafting rules, we propose to exploit the fact that current MT models generate inherently biased text: we create pseudo source segments via round-trip translation through a pivot language that (mostly) does not mark gender (Figure~\ref{fig:approach_roundtrip}). We use this method to train a gender-fair rewriter model for German, and benchmark it against a highly engineered fully rule-based baseline \citep{diesner-etal-2022-supporting}.

\subsection{Method}
\label{sec:roundtrip_method}

We propose to filter large monolingual corpora for gender-fair text \Treal and use off-the-shelve MT to first translate \Treal into a pivot language as \Ppseudo, and then translate \Ppseudo back into the original language as \Spseudo. As in \Backward, we use the resulting data to train a sequence-to-sequence model that maximises $p(\mathbf{T} | \mathbf{S}_\text{pseudo}, \boldsymbol{\theta})$.
We enrich this framework with several extensions as detailed in the next section and evaluated separately in Section~\ref{sec:roundtrip_eval_auto}.

\subsection{Experimental Setup}
\label{sec:roundtrip_experiment}

\paragraph{Data} We filter OSCAR \citep{abadji-etal-2022-towards} for German sentences that contain at least one gender-fair form with simple regular expressions that match gender-fair patterns (Appendix~\ref{app:german_patterns}). After creating pseudo sources with round-trip translation (see next paragraph) and removing duplicates and noisy segments with \texttt{OpusFilter} \citep{aulamo-etal-2020-opusfilter}, we obtain 8.8M parallel sentences. As in our \Backward experiment (Section~\ref{sec:backward_experiment}), we complement the augmented training data with copies of the gender-fair segments on the source side and non-gendered segments where no rewriting is necessary (amounting to 30\% of total data).

\paragraph{Roundtrip Translation} English is a natural choice for the pivot language since it does not express gender in most animate words, meaning that this information is often lost when translating from a language with grammatical gender to English. Indeed, we find that gender-fair forms are translated to generic masculine forms in about 90\% of the cases when we translate them to English and back to German.\footnote{As manually evaluated on a random set of 100 sentences.}  We make use of this bias to create pseudo source segments, without the need for any hand-crafted rules, by leveraging Facebook's WMT 2019 models \citep{ng-etal-2019-facebook} for German-to-English\footnote{\url{https://huggingface.co/facebook/wmt19-de-en}} and English-to-German\footnote{\url{https://huggingface.co/facebook/wmt19-en-de}}. To avoid training on other translation differences aside from gender-fair forms, we identify the counterparts of gender-fair words in the round-trip translation and merge those into the original gender-fair segment to form the pseudo source. We explain our merging algorithm in detail in Appendix \ref{app:merging}.

\paragraph{LM Prompting}

One potential issue we discovered with our training data is that gender-fair plural noun forms are much more frequent than gender-fair singular noun forms. To boost singular forms, we generate additional gender-fair training data by prompting \texttt{GerPT2-large}\footnote{\url{https://huggingface.co/benjamin/gerpt2-large}} \citep{Minixhofer_GerPT2_German_large_2020} -- a large German language model -- using a seed list of gender-fair animate nouns.\footnote{See an example prompt and verification that prompting is also possible for other languages in Appendix \ref{app:prompting}.} 
Since we do not want to bias the model towards segments that start with a prompt, we sentence split the language model outputs and only keep singular-form segments that either do not start with a prompt or that contain at least one other gender-fair form.

\paragraph{Gender Control}

As the majority of the nouns in the German round-trip outputs are masculine forms, we create additional training data by finetuning the English-to-German MT model on data marked with sentence-level gender tags (Appendix \ref{app:pair_forms}), similar to a previous approach for controlling politeness in MT outputs \citep{sennrich-etal-2016-controlling}.
We leverage the original training data\footnote{\url{https://huggingface.co/datasets/wmt19}} for the WMT 2019 shared task and finetune the original \texttt{wmt19-en-de} checkpoint for 50,000 steps with a batch size of 30 on a single GPU, following the official Hugging Face \citep{wolf-etal-2020-transformers} translation finetuning script. The resulting model does not always translate according to the given tag but produces much more balanced translations overall: with the feminine tag, only 36\% of the produced forms are masculine as compared to 90\% with the original checkpoint and 94\% with the masculine tag.\footnote{As manually evaluated on a random set of 100 sentences.} 

\paragraph{Model Architecture} We train a Transformer model using the same hyperparameters and training procedure as in our \Backward experiment described in Section~\ref{sec:backward_experiment}.

\subsection{Automatic Evaluation}
\label{sec:roundtrip_eval_auto}

We compare the performance of our model to the rule-based rewriter by \citet{diesner-etal-2022-supporting}; we are not aware of any neural rewriter for German.

\paragraph{Test Set} \citet{diesner-etal-2022-supporting} evaluate their system on a subset of the TIGER Treebank \citep{brants-etal-2004-tiger} with German news articles from the 1990s. Since masculine forms are prevalent in these articles, we create a random set of 1,200 TIGER sentences with gendered forms which we balance for masculine, feminine, singular and plural forms, and a random set of 300 non-gendered sentences. For singular forms, we decide to create our test set with a balanced mix of forms referring to unspecific people as well as real persons. There are two reasons for this design choice: First, we want to closely mirror the setup in English, where e.g.\ any occurrence of ``she'' or ``he'' is rewritten to ``they'', irrespective of whether it refers to a specific person or not. Second, there are several cases where we cannot assume that an input text referring to a specific person uses their desired pronouns. One example is machine translation output from a language that does not mark gender on pronouns. We believe that a rewriter that produces gender-fair pronouns (i.e.\ mentioning all genders) is less biased than one that assumes all gender mentions in the input text are correct (while those could be actual instances of misgendering).

\paragraph{Method} We compute tokenised WER and statistical significance as in Section~\ref{sec:backward_eval_auto}.

\paragraph{Results} Results are shown in Table \ref{tab:wer_german}. Our simplest model (\Roundtrip), which is only trained on filtered data round-tripped with the original Facebook model, already reduces the WER compared to the biased inputs from the test set (no rewriting). Avoiding differences in roundtrip translations aside from gender-fair forms (+ merged) further reduces the WER, as do additional training examples obtained through a language model (+ LM prompting) and an MT system finetuned for gender control (+ gender control).

Combining all of these extensions into a single model (+ all) results in the best WER. It also performs surprisingly well compared to \citet{diesner-etal-2022-supporting}, a rule-based system that uses an abundance of language-specific NLP tools for gender-fair rewriting (Section~\ref{sec:background_other}).

\begin{table}[]
    \centering
    \small
    \begin{tabular}{lcc}
    & \multicolumn{2}{c}{\textbf{TIGER}}  \\  \cmidrule(lr){2-3}  
         &  \textbf{non-gendered} & \textbf{gendered}  \\ \cmidrule(lr){2-2}  \cmidrule(lr){3-3} 
 Source (no rewriting) & \textbf{0.00} &  20.56\vspace{2mm}\\
    \citeauthor{diesner-etal-2022-supporting} \\\citeyearpar{diesner-etal-2022-supporting}& 0.17 & 15.01\vspace{2mm}\\
    \Roundtrip \\(this work) & 0.13 &  17.95 \\
    + merged & 0.29 & 16.36  \\
    + merged + LM prompting & 0.27 & 15.37  \\
    + merged + gender control & 0.17 & 14.02  \\
    \textbf{+ all}  & 0.17 & \textbf{13.18}
    \end{tabular}
    \caption{Tokenised WER (lower is better) of different rewriting approaches for German. Best systems (no other statistically significantly better) marked in bold.}
    \label{tab:wer_german}
\end{table}

\subsection{Human Evaluation}
\label{sec:human}
\label{sec:backward_eval_human}

While making biased text more gender-fair according to our automatic evaluation, our best-performing model still produces numerous errors: 13.18\% of the words in the model's output differ from the gender-fair reference texts in our test set (Table~\ref{tab:wer_german}). We conduct a human quality rating experiment to assess whether the imperfect outputs of this model are perceived as more gender-fair and whether erroneous gender-fair forms in general are preferred over unaltered gender-biased text.

\paragraph{Participants} To refrain from convenience sampling and to focus on potential beneficiaries of our model (i.e., people who may be offended by gender-biased text), we recruit 294 volunteers (141 female, 82 non-binary, 55 male, 16 other) through unpaid posts in newsletters and social media channels of special interest groups against gender discrimination (Appendix~\ref{app:survey_contacts}). Participation is anonymous and voluntary.

\paragraph{Materials} 
We select one paragraph each from six unrelated German texts (P1--6, Appendix~\ref{app:survey_texts}). Each paragraph contains at least one gender-biased word and is about an unspecific person whose gender is unknown, a specific non-binary person, and/or a group of people whose gender is unknown. We use three transformations of each paragraph in the experiment: unaltered (\Original), automatically de-biased by our best model (\Rewriter), and manually de-biased by one of the authors (\Gold).

\paragraph{Task and Procedure} 
We use a 6 (paragraphs) x 3 (transformations) mixed factorial design, implemented as a questionnaire in three versions (A--C) to which participants are assigned at random. Participants see all factor levels but not all combinations: to avoid repetition priming, each questionnaire includes all six paragraphs in the same order but in different transformations. For example, P1 is presented as \Original in questionnaire A and as \Rewriter in questionnaire B, etc.; participants are not informed if and how the paragraphs were transformed for gender-fairness.

After completing a pre-experiment survey with demographic questions, participants are shown a single paragraph at a time and asked if that paragraph is gender-fair, which they answer using a 5-point Likert scale (1: strongly disagree, 5: strongly agree). A short post-experiment survey on general opinions regarding gender-fair writing concludes the experiment.

\begin{figure}
    \centering
    \includegraphics[width=\columnwidth,trim=34 34 32 34,clip]{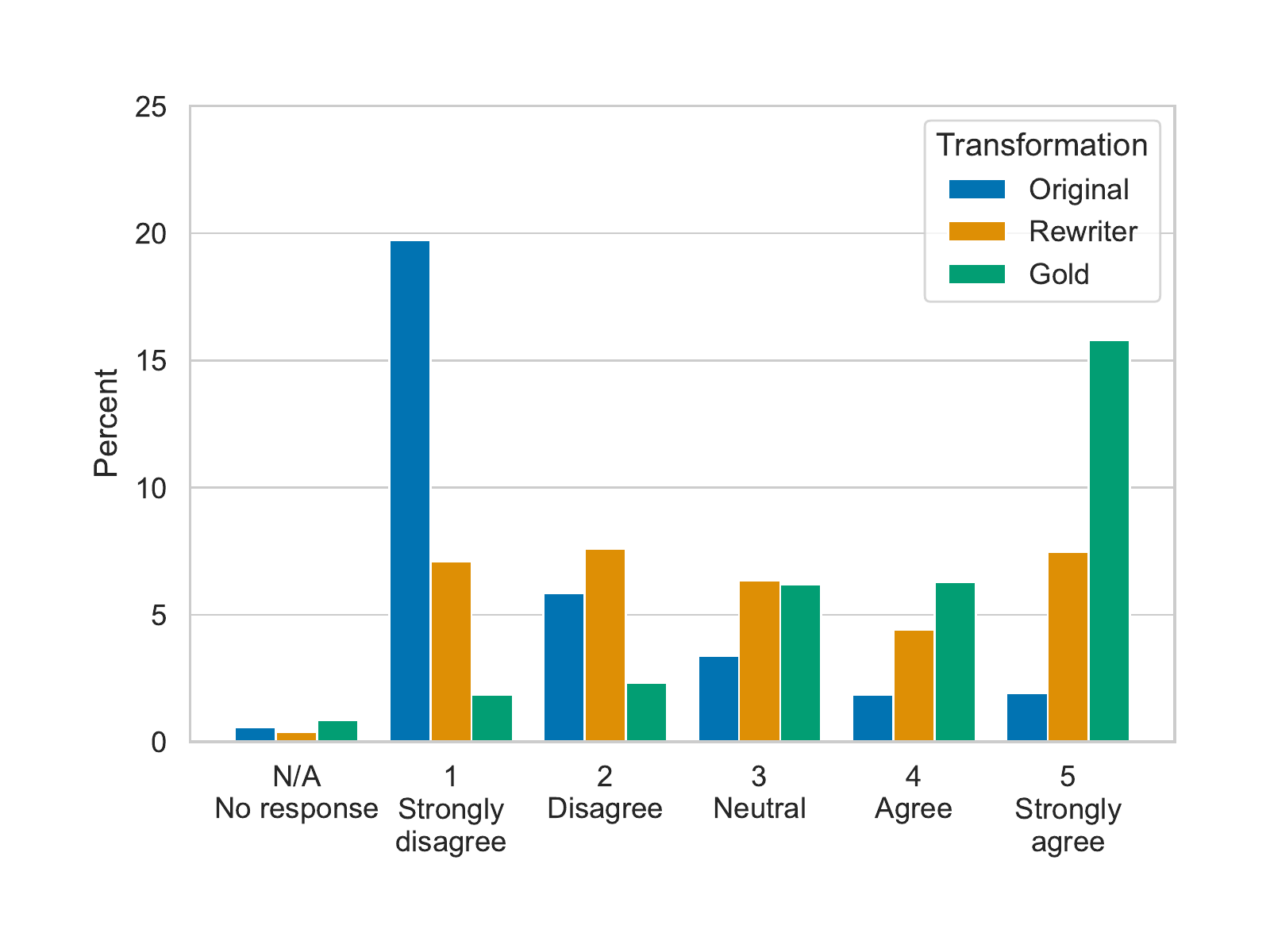}
    \caption{Distribution of responses (N=1,764) to the statement ``This text is gender-fair'' over all paragraphs.}
    \label{fig:plot_genderfair}
\end{figure}

\paragraph{Results} The distribution of Likert-scale ratings by transformation is shown in Figure~\ref{fig:plot_genderfair}. Albeit not as good as the human references (\Gold, mean=3.98), our model outputs (\Rewriter, mean=2.93) are rated better than the unaltered gender-biased paragraphs (\Original, mean=1.79) overall. This finding equally holds for all individual paragraphs (Table~\ref{tab:avg_rating_genderfair}): \Rewriter consistently outperforms \Original in terms of perceived gender-fairness and, in some cases, comes close to \Gold (P5, P6).

While these findings confirm -- as in our automatic evaluation in Section~\ref{sec:backward_eval_auto} -- that our model outputs are more gender-fair than original input texts, it leaves the most relevant question for use in practice unanswered: if given the choice, would users choose potentially erroneous \Rewriter outputs over error-free but gender-biased original texts? We include this question in the post-experiment survey independently of any specific text, and compare participants' responses with their average rating for \Original and \Rewriter outputs in the main part of the experiment. Out of the 294 participants, 201 (68.37\%) disagreed or strongly disagreed that ``If a text has errors in gender-fair wording, I prefer an error-free non-gender-fair version (e.g., generic masculine) instead.'', and 198 (98.51\%) of these participants rated \Rewriter more gender-fair than \Original in the experiment. In comparison, 35 (68.63\%) of the 51 participants who agreed or strongly agreed with that statement still gave higher gender-fairness scores to \Rewriter in the experiment (where, recall from above, the type of transformation was not revealed).

\begin{table}[]
    \small
    \setlength{\tabcolsep}{4.8pt}
    \hspace{-2mm}\begin{tabular}{llllllll}                \\
                   & P1   & P2   & P3   & P4   & P5   & P6   & All  \\ \cmidrule(lr){2-7} \cmidrule(lr){8-8}
    \Original       & 2.25 & 1.82 & 1.27 & 1.67 & 1.57 & 2.18 & 1.79 \\
    \Rewriter       & 2.97 & 3.27 & 1.86 & 2.21 & 3.56 & 3.71 & 2.93 \\
    \Gold           & 4.21 & 3.91 & 4.03 & 3.67 & 4.05 & 4.01 & 3.98
    \end{tabular}
    \caption{Mean rating on gender-fairness by paragraph (P1--6) and overall (All).}
    \label{tab:avg_rating_genderfair}
\end{table}

\section{Discussion}
\label{sec:discussion}

Our experimental findings yield insights for future research and design implications for real-world NLP applications.

\paragraph{Biased models are useful for de-biasing.}
At least in the subfield of gender-fair rewriting, de-biasing research has focussed extensively on human annotation \citep{qian-etal-2022-perturbation} and rule-based processing and training data creation \citep[e.g.,][]{sun-etal-2021-they,vanmassenhove-etal-2021-neutral,jain-etal-2021-generating, alhafni-etal-2022-user,diesner-etal-2022-supporting}. Conversely, our work demonstrates that robust de-biasing rewriters can be implemented by leveraging inherently biased NLP models.   Our \Roundtrip experiment covers a single language, but we note that both the training data \citep{abadji-etal-2022-towards} and MT models \citep{ng-etal-2019-facebook} we leverage are readily available for many other languages. We also assume that (biased) MT models -- the only requirement for gender-fair rewriters based on \Roundtrip apart from simple data filters -- are available for more languages or are easier to create than the different NLP tools used in typical rule-based augmentation pipelines, such as robust models for lemmatisation, morphological analysis, dependency parsing, named entity recognition, and co-reference resolution.

\paragraph{Rule-based de-biasing lacks robustness.}
Handwritten rules are limited by design. For example, a breakdown of the results shown in Table~\ref{tab:wer_german} (see Appendix \ref{app:detailed-results}) reveals that \citeg{diesner-etal-2022-supporting} rewriter handles masculine forms  better than our model (with a WER as low as 7.81). However, their rewriter performs poorly with feminine forms (with a WER as high as 22.22, which is worse than the WER of the biased input texts) likely because these forms are not covered in its rule set. 
Additionally, we find that while \citeg{diesner-etal-2022-supporting} approach features a solution for compounds, e.g.\ it can correctly de-bias ``Strassenbauarbeiter'' (road construction worker), this only applies to compounds where the gendered word appears at the end, e.g.\ it does not de-bias ``Arbeitergesetz'' (employee legislation). Furthermore, their approach uses a word database to identify gendered nouns which does not generalise to unknown gendered words. Finally, there is always a risk of error propagation with the NLP tools used in rule-based approaches. We conclude that language-specific rule sets will likely never cover all relevant phenomena for gender-fair rewriting. As shown by previous work, seq2seq models provide a model-based alternative that boosts generalisation \citep{vanmassenhove-etal-2021-neutral} which is why they should be used in as many languages as possible. We believe that \Roundtrip provides an easy way to create parallel data to train gender-fair rewriting models for new languages without the need for in-depth linguistic knowledge of the language.

\paragraph{Users prefer errors over bias.}
Potential beneficiaries of gender-fair rewriters -- the 294 participants of our human evaluation campaign -- rated the outputs of our German rewriter as more gender-fair than the biased original texts and explicitly (in the post-experiment survey) stated they prefer (potentially erroneous) gender-fair forms over error-free non-gender-fair forms. This is an important finding because our model is far from perfect, as evidenced by a high error rate compared to English and manual inspection of the outputs used in the evaluation campaign (Appendix~\ref{app:survey_texts}).
Previous work has found that non-binary people consider NLP as harmful, particularly due to the risk of being misgendered by MT outputs \citep{dev-etal-2021-harms}. \citet{vanmassenhove-etal-2021-neutral} caution that gender-fair rewriters may not be applicable to languages other than English because ``few languages have a crystallized approach when it comes to gender-neutral pronouns and gender-neutral word endings.'' While there is an active debate (and no established standard) about the form that gender-fair German should take \citep[e.g.,][]{burtscher-etal-2022-es}, our evaluation campaign makes a strong case for using NLP technology -- even if not flawless -- to offer \textit{one} form of de-biased text in real-world applications. Since rewriters are relatively lightweight models that operate independently from any input provider, be it a human author or a biased MT model, they would seem suitable for integration into a wide range of systems with reasonable effort.

\section{Conclusions and Future Work}
\label{sec:conclusion}

Despite an impressive performance in a wide range of tasks and applications, state-of-the-art NLP models contain numerous biases \citep{stanovsky-etal-2019-evaluating, nadeem-etal-2021-stereoset, renduchintala-williams-2022-investigating}. Our work shows that knowledge of a bias can be used to correct that bias with the biased models themselves. In the case of gender-fair rewriting, we demonstrated that reversing the data augmentation direction and using round-trip translations from biased MT models can substitute the prevalent rewriting paradigm that relies on handcrafted and often complex rules on top of morphological analysers, dependency parsers, and many other NLP tools. In our case study for German, our model surpasses the performance of a strong baseline in terms of WER and produces outputs that were perceived as more gender-fair than unaltered biased text in a human evaluation campaign with 294 potential beneficiaries.

While our approach enables the application of gender-fair rewriting to any language for which translation models exist, we believe there are several other uses cases where biased models can be leveraged for de-biasing, including dialect rewriting \citep{sun-etal-2022-dialect}, subjective bias neutralisation \citep{pryzant-2020-subjective-bias}, and avoiding discrimination in dialogue systems \citep{sheng-2021-dialogue-bias}.

\section*{Limitations}

While we consider our approach more easily applicable to new languages than rule-based \Forward, it relies on the existence of sufficient original gender-fair text in the language of interest and it is currently unclear what the minimum amount of parallel data is to learn a gender-fair rewriting model. Additionally, our survey only targets affinity groups which limits the generalisability of our results to all German speakers. Since people who choose to not use gender-fair language can simply not use a rewriting system, we do not think that this lack of generalisability is a problem in this case. Another limitation is that we use a specific form of gender-fair German in our survey. We made participants aware of this in a disclaimer at the beginning of the survey. It should be stated that there are many different acceptable gender-fair forms in German (see Section \ref{app:german_patterns}). While using a different gender-fair form could affect the individual ratings in our survey, we do not expect that it would change our finding that \Rewriter outputs are rated more gender-fair than the \Original texts.

\section*{Ethics Statement}
Participation in our study was voluntary and fully anonymous. We did not collect any personal data that would allow us to identify people and did not exclude any participants unless they specifically requested their participation be ignored in the last open commentary field of our survey or they stated technical difficulties. Concerning our rewriting models, we did not filter the publicly available data to exclude harmful content. However, since our models mainly learn to copy text, we do not believe they will hallucinate such text of their own accord.

\section*{Acknowledgements}
We thank Marcos Cramer for their valuable inputs and support with our human evaluation and all contacts who shared our survey on their mailing lists and social media accounts (Appendix \ref{app:survey_contacts}) as well as all survey participants. Many thanks also go to the \href{https://oscar-project.org/}{OSCAR project} who shared their data with us. Further, we are grateful to Alex Flückiger, Anne Göhring, Nikita Moghe and the anonymous reviewers for their helpful feedback. Chantal Amrhein and Rico Sennrich received funding from the Swiss National Science Foundation (project MUTAMUR; no. 176727).

\bibliography{anthology,custom}
\bibliographystyle{acl_natbib}

\appendix

\section{Additional Data Details}
\label{app:data-overview}

When filtering our data with \texttt{OpusFilter} \citep{aulamo-etal-2020-opusfilter}, we define noisy segments as segments that do not pass the following filters:

\begin{itemize}
    \item LengthFilter: unit=word, min=1, max=150
    \item LongWordFilter: threshold=40
    \item AlphabetRatioFilter: threshold=0.5
    \item LanguageIDFilter fasttext: threshold=0.0
\end{itemize}

The individual dataset sizes after deduplication and filtering can be seen in Table \ref{tab:data-statistics}. Note that for English, we restricted the total of gendered data to 15 million parallel segments to be comparable to \citet{sun-etal-2021-they} by only considering a subset of the English portion of the OSCAR corpus \citep{abadji-etal-2022-towards}. 

For German, we use \Roundtrip to produce the gendered pseudo sources. The finetuned MT model we use to produce feminine pseudo sources sometimes produces round-trip translations that are identical to the gender-fair segments or the round-trip translations with the original checkpoint. Consequently, the Table has fewer parallel segments for this category because fewer unseen segments are added to the training data overall for + gender control models in Table \ref{tab:wer_german}.

\begin{table}[h]
    \centering
    \small
    \begin{tabular}{cccc}
        & \multicolumn{3}{c}{\# SRC segments} \\ \cmidrule(lr){2-4}
        & masculine & feminine & gender-fair  \\ \cmidrule(lr){2-2} \cmidrule(lr){3-3} \cmidrule(lr){4-4}
        English \\ \cmidrule(lr){1-1}
       OSCAR SRC & $\sim$ 5.0M & $\sim$ 5.0M & $\sim$ 5.0M \\
       OSCAR TRG & $\sim$ 5.0M & $\sim$ 5.0M & $\sim$ 5.0M \\\\
       German  \\ \cmidrule(lr){1-1}
       OSCAR TRG & $\sim$ 8.8M & $\sim$ 8.7M & $\sim$ 8.8M  \\
       LM TRG &  $\sim$ 3.4M &  $\sim$ 2.0M &  $\sim$ 3.4M \\
    \end{tabular}
    \caption{Data statistics for all rewriter models. Gender-fair sources are copies of the target segments before normalisation. For English, masculine and feminine are created with a rule-based approach. For German, masculine refers to the round-trip translations from the original MT model checkpoint, and feminine refers to the round-trip translations from the finetuned checkpoint with the feminine tag present.}
    \label{tab:data-statistics}
\end{table}

Our English models in Table \ref{tab:wer_english} are trained on the following dataset combinations:

\begin{itemize}
    \item \textbf{\Forward Reimplementation (a+b)}: OSCAR SRC masculine + feminine + gender-fair
    \item \textbf{\Backward}: OSCAR TRG masculine + feminine + gender-fair
\end{itemize}

Our German models in Table \ref{tab:wer_german} are trained on the following dataset combinations:

\begin{itemize}
    \item \textbf{\Roundtrip (+ merged)}: OSCAR TRG masculine + gender-fair
    \item \textbf{+ LM prompting}: OSCAR TRG masculine + gender-fair and  LM TRG masculine + gender-fair
    \item \textbf{+ gender control}: OSCAR TRG masculine + feminine + gender-fair
    \item \textbf{+ all}: OSCAR TRG masculine + feminine + gender-fair and  LM TRG masculine + feminine + gender-fair
\end{itemize}

Note that for every combination of data sets, we also add non-gendered data such that the non-gendered data from OSCAR makes up 30\% of the total parallel data.

\section{Detailed Results for German}
\label{app:detailed-results}
We provide a more detailed evaluation of our results for German in Table \ref{tab:wer_german_detailed}. The first two columns show the results isolated for generic feminine and generic masculine forms in the input that should be rewritten as gender-fair forms. The third and fourth columns show the results grouped by plural and singular gender-fair forms, respectively. This evaluation highlights two points. First, we can see our strategies introduced in Section \ref{sec:roundtrip_method} are effective for the cases they were designed for: Using a gender-aware machine translation model for round-trip translation (+ gender control) is particularly helpful on the feminine test set and adding language model generated singular-form training data (+ LM prompting) reduces the WER on the singular test set significantly. Second, as discussed in Section \ref{sec:discussion}, the results show that the rule-based approach by \citet{diesner-etal-2022-supporting} is limited by design. While it performs well on generic masculine forms, it does not cover generic feminine forms at all which results in a higher WER than the source where no rewriting is performed.

\begin{table*}[]
    \centering
    \small
    \begin{tabular}{lcccc}
    & \multicolumn{4}{c}{\textbf{TIGER}}  \\  \cmidrule(lr){2-5}  
         &  \textbf{feminine} & \textbf{masculine} & \textbf{plural} & \textbf{singular}  \\ \cmidrule(lr){2-2}  \cmidrule(lr){3-3} \cmidrule(lr){4-4}  \cmidrule(lr){5-5} 
 Source (no rewriting) & 22.05 &  19.07\vspace{2mm} & 15.98 & 24.06 \\
    \citeauthor{diesner-etal-2022-supporting} \\\citeyearpar{diesner-etal-2022-supporting}& 22.22 & \textbf{\phantom{1}7.81}\vspace{2mm} & 10.39 & \textbf{18.55} \\
    \Roundtrip \\(this work) & 19.76 &  16.13 & 11.04 & 23.22 \\
    + merged & 18.99 & 13.72 & \phantom{1}8.29 & 22.52 \\
    + merged + LM prompting & 17.51 & 13.24 & \phantom{1}8.05 & 20.97\\
    + merged + gender control & 14.00 & 14.04 & \textbf{\phantom{1}4.12} & 21.58 \\
    \textbf{+ all}  & \textbf{13.03} & 13.33 & \phantom{1}4.67 & 19.69
    \end{tabular}
    \caption{Tokenised WER (lower is better) of different rewriting approaches for German evaluated separately for different test cases. Best systems (no other statistically significantly better) marked in bold.}
    \label{tab:wer_german_detailed}
\end{table*}

\section{Merging Algorithm}
\label{app:merging}
One issue with round-trip translations is that they are likely to contain edits unrelated to the gender-fair words in the target:\\

\begin{small}
\begin{tabularx}{0.95\columnwidth}{rX}
    original (de) &  Denn jede Begegnung mit einem*r Schüler*in ist anders, auch die Familien sind verschieden.\\\\
    interm. (en) & Because every encounter with a student is different, even the families are different. \\\\
    round-trip (de) & \textbf{Weil} jede Begegnung mit \textcolor{highlight}{\textbf{einem Schüler}} \textbf{anders ist, sind} auch die Familien \textbf{anders}.
\end{tabularx}
\end{small}

This round-trip translation not only contains the desired generic masculine form for student (marked in orange) but also several other deviations from the original sentence (marked in bold) which even changes the meaning slightly.

Ideally, we would want to identify the generic masculine form and allow no other changes in the pseudo source:\\

\begin{small}
\begin{tabularx}{0.95\columnwidth}{rX}
    merged (de) & Denn jede Begegnung mit \textcolor{highlight}{\textbf{einem Schüler}} ist anders, auch die Familien sind verschieden.
\end{tabularx}
\end{small}

To this end, we develop a merging algorithm that aims to insert the generic forms in the round-trip translation into the context of the original gender-fair segment. For all gender-fair words in the target segment, we check if there are any close matches in the round-trip translation using the \texttt{difflib} Python library\footnote{\url{https://docs.python.org/3/library/difflib.html\#difflib.get_close_matches}} with a cutoff of 0.6. If yes, we replace the gender-fair word with its closest match. This process can be seen in Algorithm \ref{algo:merging}. If not all gender-fair words can be matched in the round-trip translation, we keep the round-trip translation as our pseudo source and do not merge with the gender-fair target. Note that this merging algorithm can potentially introduce case and other grammatical errors in the pseudo source. This is, however, not a serious problem as these potentially non-grammatical forms will only occur on the source side, meaning our model does not learn to produce such forms on the target side. The merging algorithm could be improved in the future to also consider grammatical acceptability, e.g.\ by scoring the merged pseudo source against the gender-fair target with a language model.

\begin{algorithm}[t]
\caption{Merging Round-Trip Translations}
\label{algo:merging}
\begin{algorithmic}[1]
\Require list of tokens in gender-fair target $t$, list of tokens in RT translation $r$
\Ensure merged pseudo source $s$
\For{token $w$ at index $i$ in $t$}
        \If{$w$ is gender-fair form} 
            \State $m$ =  get\_close\_matches($w$, $r$, 0.6)
            \If{len($m$) $> 0$} 
                $t[i]$ = $m[0]$
            \EndIf
        \EndIf
      \EndFor
\State $s$ = detokenise($t$)
\end{algorithmic}
\end{algorithm}

\section{Additional LM Prompting Details}
\label{app:prompting}

Our prompts consist of a gender-fair determiner and noun from a seed list, following the pattern ``Ein*e \texttt{NOUN}*in''. An example can be seen here with the prompt in bold and gender-fair forms generated by the language model in orange:\\

German:
\begin{small}
\textbf{Ein*e Leiter*in} für unser Team „Soziales“.
Als \textcolor{highlight}{Ansprechpartner*in} in Sachen Finanzen oder für Veranstaltungen, mit einem Blick für’s Wesentliche sind Sie für uns eine Bereicherung
Wir suchen \textcolor{highlight}{eine*n freundliche*n, zuverlässige*n} und \textcolor{highlight}{flexible*n Mitarbeiter*in} für eine geringfügige Beschäftigung. Sie haben Freude am Umgang mit \textcolor{highlight}{Kolleg*innen} und eine hohe Einsatzbereitschaft. Auch wenn Sie noch keine Erfahrung mitbringen lohnt sich dies schon sehr. Eine abgeschlossene Ausbildung als \textcolor{highlight}{Köch*in} oder eine vergleichbare Qualifikation sind von Vorteil. Es erwartet Sie ein spannendes Arbeitsumfeld und ein gutes Betriebsklima sowie verschiedene Weiterbildungs- und Fortbildungsmöglichkeiten.
\end{small}\\

Prompting large language models for gender-fair text also works with other language models and for languages other than German. Here, we show example gender-fair prompts (marked in bold) and the text generated by a multilingual language model\footnote{\url{https://huggingface.co/bigscience/bloom}}. Many forms in the generated text are also gender-fair (marked in orange) even over long distances. Not all generated forms include non-binary people (e.g.\ French ``il/elle'') but these can easily be normalised.\\

French:
\begin{small}
\textbf{Un·e citoyen·ne} parmi tant d’autres, qui vit sur la commune et la fréquente dans toute sa diversité. \textcolor{highlight}{L’un·e} qui a envie de vous donner de son temps pour la ville dans laquelle \textcolor{highlight}{il/elle} vit. C’est avant tout une adhésion à notre vision d’une ville plus durable, plus juste, plus proche de vous. En devenant \textcolor{highlight}{membre}, vous agissez directement en faveur de la démocratie locale, parce que \textcolor{highlight}{les citoyen·nes} comme vous peuvent exprimer leur avis, ou défendre les valeurs qu’\textcolor{highlight}{ils et elles} partagent et portent avec nous. Et vous pouvez le faire par vos votes aux assemblées générales, ou dans le cadre des groupes thématiques dans lesquels vous \textcolor{highlight}{êtes invité·e} à vous engager.
\end{small}\\

Spanish:
\begin{small}
\textbf{Unx companerx} \textcolor{highlight}{presx}. Esa son nuestras armas, pensadlo. Vayamos al Estado con la fuerza de todas nuestras rebeliones, hacernos fuertes en todo y donde \textcolor{highlight}{todxs} podamos participar, unificándonos como un movimiento, concentrando la rabia y lanzándola contra quienes nos mantienen en la miseria, que eso es lo que hacen estos gobiernos a base de recortes, no trabajan para el pueblo, se han posicionado en contra de \textcolor{highlight}{lxs más desprotegidxs}, de \textcolor{highlight}{lxs desempleadxs}, de \textcolor{highlight}{lxs jóvenes}, de \textcolor{highlight}{lxs de mediana edad}. Contra \textcolor{highlight}{todxs nosotrxs}, contra \textcolor{highlight}{lxs que resistimos} al poder y luchamos en sus filas de igual a igual, contra \textcolor{highlight}{lxs de arriba y abajo}, contra el capitalismo y el Estado y su falsa democracia.
\end{small}

\section{English Lookup Tables}
\label{app:english_lookup}

For completeness, we list the lookup tables used in our reproduction of \citet{vanmassenhove-etal-2021-neutral} and \citet{sun-etal-2021-they} and in the other direction for our \Backward.

\subsection{Pronouns}
\label{app:english_lookup_pronouns}
\begin{center}
\small
\vspace{0.5cm}
\begin{tabular}{ccc}
    he, she & $\leftrightarrow$ & they\\
    his, her & $\leftrightarrow$ & their\\
    him, her & $\leftrightarrow$ & them\\
    his, hers & $\leftrightarrow$ & theirs\\
    himself, herself & $\leftrightarrow$ & themself\\
\end{tabular}
\end{center}

\subsection{Nouns}
\label{app:english_lookup_nouns}

\begin{center}
\tiny
\textbf{Gender-neutral alternatives for gender-marked job titles}\\
\vspace{0.2cm}
    \begin{tabular}{ccc}
        chairman, chairwoman & $\leftrightarrow$ & chairperson \\
        chairmen, chairwomen & $\leftrightarrow$ & chairpeople \\
        anchorman, anchorwoman & $\leftrightarrow$ & anchor \\
        anchormen, anchorwomen & $\leftrightarrow$ & anchors \\
        congressman, congresswoman & $\leftrightarrow$ & member of congress \\
        congressmen, congresswomen & $\leftrightarrow$ & members of congress \\
        policeman, policewoman & $\leftrightarrow$ & police officer \\
        policemen, policewomen & $\leftrightarrow$ & police officers \\
        spokesman, spokeswoman & $\leftrightarrow$ & spokesperson \\
        spokesmen, spokeswomen & $\leftrightarrow$ & spokespeople \\
        steward, stewardess & $\leftrightarrow$ & flight attendant \\
        stewards, stewardesses & $\leftrightarrow$ & flight attendants \\
        headmaster, headmistress & $\leftrightarrow$ & principal \\
        headmasters, headmistresses & $\leftrightarrow$ & principals \\
        businessman, businesswoman & $\leftrightarrow$ & business person \\
        businessmen, businesswomen & $\leftrightarrow$ & business persons \\
        postman, postwoman & $\leftrightarrow$ & mail carrier \\
        postmen, postwomen & $\leftrightarrow$ & mail carriers \\
        salesman, saleswoman & $\leftrightarrow$ & salesperson \\
        salesmen, saleswomen & $\leftrightarrow$ & salespersons \\
        fireman, firewoman & $\leftrightarrow$ & firefighter \\
        firemen, firewomen & $\leftrightarrow$ & firefighters \\
        barman, barwoman & $\leftrightarrow$ & bartender \\
        barmen, barwomen & $\leftrightarrow$ & bartenders \\
        cleaning man, cleaning lady & $\leftrightarrow$ & cleaner \\
        cleaning men, cleaning ladies & $\leftrightarrow$ & cleaners \\
        foreman, forewoman & $\leftrightarrow$ & supervisor \\
        foremen, forewomen & $\leftrightarrow$ & supervisors 
    \end{tabular}
\end{center}

\begin{center}
\tiny
\textbf{Gender-neutral alternatives for generic ‘man’}\\
\vspace{0.2cm}
    \begin{tabular}{ccc}
        average man & $\leftrightarrow$ & average person \\
        average men & $\leftrightarrow$ & average people \\
        best man for the job & $\leftrightarrow$ & best person for the job \\
        best men for the job & $\leftrightarrow$ & best people for the job \\
        layman & $\leftrightarrow$ & layperson \\
        laymen & $\leftrightarrow$ & laypeople \\
        man and wife & $\leftrightarrow$ & husband and wife \\
        mankind & $\leftrightarrow$ & humankind \\
        man-made & $\leftrightarrow$ & human-made \\
        workmanlike & $\leftrightarrow$ & skillful \\
        freshman & $\leftrightarrow$ & first-year student
\end{tabular}
\end{center}    

\begin{center}
\tiny
\textbf{Gender-neutral alternatives for unnecessary feminine forms}\\
\vspace{0.2cm}
    \begin{tabular}{ccc}
        actress & $\leftrightarrow$ & actor \\
        actresses & $\leftrightarrow$ & actors \\
        heroine & $\leftrightarrow$ & hero \\
        heroines & $\leftrightarrow$ & heroes \\
        comedienne & $\leftrightarrow$ & comedian \\
        comediennes & $\leftrightarrow$ & comedians \\
        executrix & $\leftrightarrow$ & executor \\
        executrices & $\leftrightarrow$ & executors \\
        poetess & $\leftrightarrow$ & poet \\
        poetesses & $\leftrightarrow$ & poets \\
        usherette & $\leftrightarrow$ & usher \\
        usherettes & $\leftrightarrow$ & ushers \\
        authoress & $\leftrightarrow$ & author \\
        authoresses & $\leftrightarrow$ & authors \\
        boss lady & $\leftrightarrow$ & boss \\
        boss ladies & $\leftrightarrow$ & bosses \\
        waitress & $\leftrightarrow$ & waiter \\
        waitresses & $\leftrightarrow$ & waiters 
    \end{tabular}
\end{center}

\section{German Gender-fair Patterns}
\label{app:german_patterns}

For German, we cannot use a lookup approach to identify gender-fair noun forms but rather work with several gender-fair patterns. Here, we describe the different gender-fair forms we consider and, for each, show a plural form example and its corresponding pattern:

\textbf{Pair forms} are forms that explicitly state the feminine and masculine form connected with a coordinating conjunction like ``and'' or ``or''. The order of the feminine and masculine forms can be variable. This form assumes binary gender and does not include non-binary people.

\begin{itemize}
    \item[]  Example: \hfill Studentinnen und Studenten 
    \item[]  Pattern: \hfill \small{(\textbackslash S\{2,\})innen und -?\textbackslash1(?!innen)(en|e|n)?}
\end{itemize}

\textbf{Binnen-I forms} are forms that take the feminine form but with a capitalised ``I'' at the beginning of the feminine suffix ``innen'' (plural) or ``in'' (singular). This form also assumes binary gender.

\begin{itemize}
    \item[]  Example: \hfill StudentInnen
    \item[]  Pattern: \hfill \textbackslash w+Innen
\end{itemize}

\textbf{Gender slash forms} are forms that take the feminine form but with a slash (``/'') separating the feminine suffix ``innen'' (plural) or ``in'' (singular) from the stem. This form also assumes binary gender.

\begin{itemize}
    \item[]  Example: \hfill Student/innen
    \item[]  Pattern: \hfill \textbackslash w+\textbackslash\ ?/\textbackslash\ ?innen
\end{itemize}

\textbf{Gender gap forms} are forms that take the feminine form but with an underscore (``\_'') separating the feminine suffix ``innen'' (plural) or ``in'' (singular) from the stem. This form includes non-binary people.

\begin{itemize}
    \item[]  Example: \hfill Student\_innen
    \item[]  Pattern: \hfill \textbackslash w+\_innen
\end{itemize}

\textbf{Gender colon forms} are forms that take the feminine form but with a colon (``:'') separating the feminine suffix ``innen'' (plural) or ``in'' (singular) from the stem. This form also includes non-binary people.

\begin{itemize}
    \item[]  Example: \hfill Student:innen
    \item[]  Pattern: \hfill \textbackslash w+:innen
\end{itemize}

\textbf{Gender star forms} are forms that take the feminine form but with an asterisk (``*'') separating the feminine suffix ``innen'' (plural) or ``in'' (singular) from the stem. This form also includes non-binary people.

\begin{itemize}
    \item[]  Example: \hfill Student*innen
    \item[]  Pattern: \hfill \textbackslash w+\textbackslash*innen
\end{itemize}

Alternatively, and out of the scope of this work, gender-fair text can also use present participles as gender-neutral nouns (only gender-neutral in plural forms, e.g.\ Studierende - ``those who are studying''), synonymous gender-neutral nouns or it can completely avoid gendered words and express content with structures where no gender-fair forms are needed (e.g.\ ``bei den Dorfbewohner*innen'' - ``among the villagers'' could also be expressed as ``im Dorf'' - ``in the village''). We believe that our proposed approach can be extended to those cases in the future.

\section{Gender-Tagged Data With Pair Forms}
\label{app:pair_forms}

To make the English-to-German machine translation model that we use for round-trip translations gender-aware, we finetune on artificial data with sentence-level gender labels. Previous work presented several approaches how parallel data can be filtered or created to specifically contain masculine or feminine forms \citep{costa-jussa-de-jorge-2020-fine, saunders-byrne-2020-reducing, choubey-etal-2021-gfst,corral-saralegi-2022-gender}. In our work, we create such data by making use of pair forms (see Appendix \ref{app:german_patterns}) in existing parallel data that consist of the feminine and the masculine form of the same noun:

\begin{small}
\vspace{0.5cm}
\begin{tabularx}{0.95\columnwidth}{X}
    SRC: \textbf{Students} from many nations learn together here. \\\\
    TRG: \textbf{Schülerinnen und Schüler} aus vielen Nationen lernen hier gemeinsam.
\end{tabularx}
\vspace{0.3cm}
\end{small}

Using a simple replace operation, we can use this data to create two contrasting targets in German and tag them with a corresponding tag that indicates whether the translation should contain feminine or masculine noun forms:

\begin{small}
\vspace{0.5cm}
\begin{tabularx}{0.95\columnwidth}{X}
    SRC: \textbf{<f> Students} from many nations learn together here.\\\\
    TRG: \textbf{Schülerinnen} aus vielen Nationen lernen hier gemeinsam. \\\\\\
    SRC: \textbf{<m> Students} from many nations learn together here.\\\\
    TRG: \textbf{Schüler} aus vielen Nationen lernen hier gemeinsam.
\end{tabularx}
\vspace{0.3cm}
\end{small}

This is possible for plural nouns because all German plural nouns share inflexion across genders which means that no rewriting is necessary for pronouns, adjectives or determiners that refer to them. For singular forms, we cannot easily construct contrasting versions because sometimes additional modifications to pronouns, adjectives or determiners are necessary to preserve the grammatical agreement. Instead, we create feminine examples for singular pair forms if the first form in the pair form is feminine and we create masculine examples if the first form in the pair form is masculine.

Pair forms are not only specific to German but are also common in many other languages. For example, the United Nations advise the use of pair forms in all their official languages with grammatical gender - \href{https://www.un.org/ar/gender-inclusive-language/guidelines.shtml}{Arabic}, \href{https://www.un.org/fr/gender-inclusive-language/guidelines.shtml}{French}, \href{https://www.un.org/ru/gender-inclusive-language/guidelines.shtml}{Russian} and \href{https://www.un.org/er/gender-inclusive-language/guidelines.shtml}{Spanish} - as well as in \href{https://www.un.org/en/gender-inclusive-language/guidelines.shtml}{English} and \href{https://www.un.org/zh/gender-inclusive-language/guidelines.shtml}{Chinese} for added emphasis when gender is relevant for communication. Our approach to generating finetuning data for gender-aware machine translation models based on pair forms is not limited to German but is applicable to other languages as well.

\section{Overview of Survey Texts}
\label{app:survey_texts}
The six text excerpts used in our survey and the corresponding three versions can be seen in Table \ref{tab:survey_texts}.

P1 is from a website informing about \href{https://www.odawald.ch/weiterbildung/foerster-hf/}{study regulations for becoming a forester} and uses generic masculine forms and gender slash forms (that do not include non-binary people) in plural and singular. P2 is from an \href{https://www.ruv.at/datenschutzerklaerung}{online privacy policy} of an insurance company and uses generic masculine forms in plural and singular. P3 is from \href{https://www.uno-kartenspiel.de/spielregeln/}{game instructions} and uses generic masculine in singular. P4 is from a \href{https://www.stern.de/lifestyle/leute/ezra-miller--star-aus--the-flash--erneut-festgenommen-31792692.html}{news article} about Ezra Miller who uses ``they/them'' pronouns in English and the text uses generic masculine forms in singular. P5 is from a \href{https://www.clockodo.com/de/ratgeber/17-einfache-massnahmen-zur-kundenbindung-die-wirklich-wirken/#c4594-headline}{company blog} and uses generic masculine forms in plural and singular. P6 is from a \href{https://www.exeltis.de/wp-content/uploads/2020/12/Gebrauchsinformation-Asumate-20-217.pdf}{package insert} of a birth control pill and uses generic masculine forms in singular and generic feminine forms in plural. (Websites last accessed on 27.12.2022)

\begin{table*}[]
    \centering
    \tiny
    \begin{tabularx}{\textwidth}{lXXX}
        & \textbf{Original Text} & \textbf{Rewriter Output} & \textbf{Human Reference} \\\\
        \textbf{P1} & Zugelassen zur Försterausbildung werden Kandidaten mit einem eidg. Fähigkeitszeugnis als Forstwart/in (oder einer gleichwertigen Ausbildung). Erforderlich sind zudem 12 Monate Praxis in einem Forstbetrieb oder -unternehmen. Zudem müssen Interessenten die nachfolgenden Grundlagenmodule und die Eignungsprüfung absolviert und bestanden haben. Details dazu sind erhältlich bei den Anbietern: BZW Lyss und ibW BZW Maienfeld. Angehende Forstwart-Vorarbeiter/innen und Förster/innen besuchen die gleichen Grundlagenmodule. & Zugelassen zur Försterausbildung werden Kandidaten mit einem eidg. Fähigkeitszeugnis als \textcolor{highlight}{Forstwart*in} (oder einer gleichwertigen Ausbildung). Erforderlich sind zudem 12 Monate Praxis in einem Forstbetrieb oder -unternehmen. Zudem müssen \textcolor{highlight}{Interessent*innen} die nachfolgenden Grundlagenmodule und die Eignungsprüfung absolviert und bestanden haben. Details dazu sind erhältlich bei den Anbietern: BZW Lyss und ibW BZW Maienfeld. Angehende Forstwart-\textcolor{highlight}{Vorarbeiter*innen} und \textcolor{highlight}{Förster*innen} besuchen die gleichen Grundlagenmodule. & Zugelassen zur \textcolor{highlight}{Förster*innenausbildung} werden \textcolor{highlight}{Kandidat*innen} mit einem eidg. Fähigkeitszeugnis als \textcolor{highlight}{Forstwart*in} (oder einer gleichwertigen Ausbildung). Erforderlich sind zudem 12 Monate Praxis in einem Forstbetrieb oder -unternehmen. Zudem müssen \textcolor{highlight}{Interessent*innen} die nachfolgenden Grundlagenmodule und die Eignungsprüfung absolviert und bestanden haben. Details dazu sind erhältlich bei den \textcolor{highlight}{Anbieter*innen}: BZW Lyss und ibW BZW Maienfeld. Angehende \textcolor{highlight}{Forstwart*in-Vorarbeiter*innen} und \textcolor{highlight}{Förster*innen} besuchen die gleichen Grundlagenmodule.\\\\
        \textbf{P2} &  1. Einführung

        Durch die Technik des Internets und der elektronischen Datenverarbeitung kann der Einzelne das Gefühl bekommen, den Überblick darüber zu verlieren, wo und zu welchem Zweck seine Daten gespeichert werden. Gerade im finanziellen Bereich ist das Vertrauen in die sorgfältige und sichere Behandlung von Kundendaten besonders wichtig. Deshalb möchten wir Ihnen als Besucher unserer Web-Seiten erläutern, wie die Unternehmen der R+V Versicherungsgruppe die Vertraulichkeit Ihrer personenbezogenen Daten sicherstellt und die Persönlichkeitsrechte respektiert. &
        1. Einführung

        Durch die Technik des Internets und der elektronischen Datenverarbeitung kann der Einzelne das Gefühl bekommen, den Überblick darüber zu verlieren, wo und zu welchem Zweck seine Daten gespeichert werden. Gerade im finanziellen Bereich ist das Vertrauen in die sorgfältige und sichere Behandlung von \textcolor{highlight}{Kund*innendaten} besonders wichtig. Deshalb möchten wir Ihnen als \textcolor{highlight}{Besucher*in} unserer Web-Seiten erläutern, wie die Unternehmen der R+V Versicherungsgruppe die Vertraulichkeit Ihrer personenbezogenen Daten sicherstellt und die Persönlichkeitsrechte respektiert.
        &
        1. Einführung

        Durch die Technik des Internets und der elektronischen Datenverarbeitung kann \textcolor{highlight}{der*die} Einzelne das Gefühl bekommen, den Überblick darüber zu verlieren, wo und zu welchem Zweck \textcolor{highlight}{seine*ihre} Daten gespeichert werden. Gerade im finanziellen Bereich ist das Vertrauen in die sorgfältige und sichere Behandlung von \textcolor{highlight}{Kund*innendaten} besonders wichtig. Deshalb möchten wir Ihnen als \textcolor{highlight}{Besucher*in} unserer Web-Seiten erläutern, wie die Unternehmen der R+V Versicherungsgruppe die Vertraulichkeit Ihrer personenbezogenen Daten sicherstellt und die Persönlichkeitsrechte respektiert. \\\\\\
        \textbf{P3} & Hat er keine passende Karte ist der nächste Spieler an der Reihe. Wer die vorletzte Karte ablegt, muss „UNO!“ (das bedeutet „Eins“) rufen und signalisiert damit, dass er nur noch eine Karte auf der Hand hat. Vergisst ein Spieler das und ein anderer bekommt es rechtzeitig mit (bevor der nächste Spieler eine Karte gezogen oder abgeworfen hat) so muss er 2 Strafkarten ziehen. Die Runde gewinnt derjenige, welcher die letzte Karte abgelegt hat. Die Punkte werden addiert und eine neue Runde wird gespielt. &
        Hat er keine passende Karte ist der nächste \textcolor{highlight}{Spieler*in} an der Reihe. Wer die vorletzte Karte ablegt, muss „UNO!“ (das bedeutet „Eins“) rufen und signalisiert damit, dass er nur noch eine Karte auf der Hand hat. Vergisst ein Spieler das und ein anderer bekommt es rechtzeitig mit (bevor der nächste \textcolor{highlight}{Spieler*innen} eine Karte gezogen oder abgeworfen hat) so muss er 2 Strafkarten ziehen. Die Runde gewinnt derjenige, welcher die letzte Karte abgelegt hat. Die Punkte werden addiert und eine neue Runde wird gespielt. & 
        Hat \textcolor{highlight}{er*sie} keine passende Karte ist \textcolor{highlight}{der*die} nächste \textcolor{highlight}{Spieler*in} an der Reihe. Wer die vorletzte Karte ablegt, muss „UNO!“ (das bedeutet „Eins“) rufen und signalisiert damit, dass \textcolor{highlight}{er*sie} nur noch eine Karte auf der Hand hat. Vergisst \textcolor{highlight}{ein*e Spieler*in} das und \textcolor{highlight}{ein*e andere*r} bekommt es rechtzeitig mit (bevor \textcolor{highlight}{der*die} nächste \textcolor{highlight}{Spieler*in} eine Karte gezogen oder abgeworfen hat) so muss \textcolor{highlight}{er*sie} 2 Strafkarten ziehen. Die Runde gewinnt \textcolor{highlight}{der*diejenige, welche*r} die letzte Karte abgelegt hat. Die Punkte werden addiert und eine neue Runde wird gespielt. \\\\
        \textbf{P4} & Erst Ende März war Miller bereits negativ aufgefallen. Ebenfalls auf Hawaii randalierte er in einer Karaoke-Bar in Honolulu derartig, dass die Polizei einschreiten musste und den Star wegen Ruhestörung und Belästigung festnahm. Er habe Obszönitäten geschrien und versucht, einer 23-jährigen Besucherin das Mikrofon aus der Hand zu reißen. Dies stellte laut Polizei eine Ordnungswidrigkeit dar. Später griff er einen 32 Jahre alten Mann an, der Darts spielte. Dies erfülle den Tatbestand der Belästigung.

        Seit 2016 verkörpert Miller, der sich als non-binäre Person identifiziert, im DC Extended Universe den blitzschnellen Superhelden The Flash. & Erst Ende März war Miller bereits negativ aufgefallen. Ebenfalls auf Hawaii randalierte er in einer Karaoke-Bar in Honolulu derartig, dass die Polizei einschreiten musste und den Star wegen Ruhestörung und Belästigung festnahm. Er habe Obszönitäten geschrien und versucht, einer 23-jährigen \textcolor{highlight}{Besucher*in} das Mikrofon aus der Hand zu reißen. Dies stellte laut Polizei eine Ordnungswidrigkeit dar. Später griff \textcolor{highlight}{er*sie} einen 32 Jahre alten Mann an, der Darts spielte. Dies erfülle den Tatbestand der Belästigung.

        Seit 2016 verkörpert Miller, der sich als non-binäre Person identifiziert, im DC Extended Universe den blitzschnellen Superhelden The Flash. & Erst Ende März war Miller bereits negativ aufgefallen. Ebenfalls auf Hawaii randalierte \textcolor{highlight}{er*sie} in einer Karaoke-Bar in Honolulu derartig, dass die Polizei einschreiten musste und den Star wegen Ruhestörung und Belästigung festnahm. \textcolor{highlight}{Er*sie} habe Obszönitäten geschrien und versucht, \textcolor{highlight}{einem*r} 23-jährigen \textcolor{highlight}{Besucher*in} das Mikrofon aus der Hand zu reißen. Dies stellte laut Polizei eine Ordnungswidrigkeit dar. Später griff \textcolor{highlight}{er*sie} einen 32 Jahre alten Mann an, der Darts spielte. Dies erfülle den Tatbestand der Belästigung.

        Seit 2016 verkörpert Miller, \textcolor{highlight}{der*die} sich als non-binäre Person identifiziert, im DC Extended Universe \textcolor{highlight}{den*die blitzschnelle*n Superheld*in} The Flash.\\\\\\
        \textbf{P5} &  Klassische Maßnahmen zur Kundenbindung 

        1. Kundenclub

        Durch eine Kundenkarte bekommen Käufer zum Beispiel Prozente oder andere Vorzüge, während Sie als Unternehmer die Daten erhalten. Dadurch sind Sie in der Lage, Kontakt mit ihm aufzunehmen. Es gibt kostenlose Kundenkarten und kostenpflichtige, die einen höheren Rabatt geben. &
        Klassische Maßnahmen zur Kundenbindung

        1. Kundenclub

        Durch eine Kundenkarte bekommen \textcolor{highlight}{Käufer*innen} zum Beispiel Prozente oder andere Vorzüge, während Sie als \textcolor{highlight}{Unternehmer*in} die Daten erhalten. Dadurch sind Sie in der Lage, Kontakt mit \textcolor{highlight}{ihm*ihr} aufzunehmen. Es gibt kostenlose Kundenkarten und kostenpflichtige, die einen höheren Rabatt geben. &
        Klassische Maßnahmen zur \textcolor{highlight}{Kund*innenbindung}
        
        1. \textcolor{highlight}{Kund*innenclub}

        Durch eine \textcolor{highlight}{Kund*innenkarte} bekommen \textcolor{highlight}{Käufer*innen} zum Beispiel Prozente oder andere Vorzüge, während Sie als \textcolor{highlight}{Unternehmer*in} die Daten erhalten. Dadurch sind Sie in der Lage, Kontakt mit \textcolor{highlight}{ihm*ihr} aufzunehmen. Es gibt kostenlose \textcolor{highlight}{Kund*innenkarten} und kostenpflichtige, die einen höheren Rabatt geben. \\\\\\
        \textbf{P6} & Es ist wichtig, regelmäßig Ihre Brüste untersuchen zu lassen, und Sie sollten Ihren Arzt aufsuchen, wenn Sie einen Knoten fühlen. In seltenen Fällen wurden gutartige Lebertumore und noch seltener bösartige Lebertumore bei Anwenderinnen von KOKs berichtet. Suchen Sie Ihren Arzt auf, wenn Sie ungewöhnlich starke Bauchschmerzen haben.

        Gebärmutterhalskrebs wurde bei Langzeitanwenderinnen beobachtet; aber es ist nicht geklärt, in wie weit unterschiedliches Sexualverhalten oder andere Faktoren wie das humane Papilloma-Virus (HPV) dazu beitragen. &
        Es ist wichtig, regelmäßig Ihre Brüste untersuchen zu lassen, und Sie sollten Ihren Arzt aufsuchen, wenn Sie einen Knoten fühlen. In seltenen Fällen wurden gutartige Lebertumore und noch seltener bösartige Lebertumore bei \textcolor{highlight}{Anwender*innen} von KOKs berichtet. Suchen Sie Ihren Arzt auf, wenn Sie ungewöhnlich starke Bauchschmerzen haben.

        Gebärmutterhalskrebs wurde bei \textcolor{highlight}{Langzeitanwender*innen} beobachtet; aber es ist nicht geklärt, in wie weit unterschiedliches Sexualverhalten oder andere Faktoren wie das humane Papilloma-Virus (HPV) dazu beitragen. &
        Es ist wichtig, regelmäßig Ihre Brüste untersuchen zu lassen, und Sie sollten \textcolor{highlight}{Ihre*n Ärzt*in} aufsuchen, wenn Sie einen Knoten fühlen. In seltenen Fällen wurden gutartige Lebertumore und noch seltener bösartige Lebertumore bei \textcolor{highlight}{Anwender*innen} von KOKs berichtet. Suchen Sie \textcolor{highlight}{Ihre*n Ärzt*in} auf, wenn Sie ungewöhnlich starke Bauchschmerzen haben.

        Gebärmutterhalskrebs wurde bei \textcolor{highlight}{Langzeitanwender*innen} beobachtet; aber es ist nicht geklärt, in wie weit unterschiedliches Sexualverhalten oder andere Faktoren wie das humane Papilloma-Virus (HPV) dazu beitragen.\\
    \end{tabularx}
    \caption{Different versions of six text excerpts used in our survey. Changes compared to the original text are marked in orange.}
    \label{tab:survey_texts}
\end{table*}

\section{List of Survey Contacts}
\label{app:survey_contacts}

Our survey was kindly shared on mailing lists or other platforms by the networks below.\\

Austria:
\begin{itemize}
    \item \href{https://venib.at/}{Venib - Verein Nicht-Binär}\\
\end{itemize}

Germany:
\begin{itemize}
    \item \href{https://www.facebook.com/minas.offiziell}{MinaS - Verein für Menschen im nichtbinären und agender Spektrum}
    \item \href{https://geschlechtsneutral.net/}{Verein für geschlechtsneutrales Deutsch e.V.}\\
\end{itemize}

Switzerland:
\begin{itemize}
    \item \href{https://gendercampus.ch/}{Gender Campus}
    \item \href{https://nonbinary.ch}{nonbinary.ch}
    \item \href{https://www.instagram.com/queerstudentsbern}{Queerstudents Bern}
    \item \href{https://www.instagram.com/queerkaff_ow}{Queers usem Kaff}
    \item \href{https://romanescos.ch/}{romanescos mailing list}
    \item \href{https://tgns.ch}{Transgender Network Switzerland}
    \item \href{https://geschlechtergerechter.ch/}{Verein Geschlechtergerechter}
    \item \href{https://www.instagram.com/zurichpride}{Zurich Pride}\\
\end{itemize}

Each contact was provided with a link to the survey and the following accompanying text (English translation below):\\

German original:\\\\
\begin{tiny}
\vspace{0.3cm}
\begin{tabularx}{0.95\columnwidth}{X}
\texttt{``Das Institut für Computerlinguistik (Uni Zürich) und Textshuttle arbeiten an Textgenerierungssystemen, die genderfair(er)e Sprache ausgeben und möchten in einer Umfrage untersuchen, wie unterschiedliche Texte in Bezug auf genderfaire Sprache wahrgenommen werden.}\\\\
\texttt{Die Umfrage dauert 10–15 Minuten und ist anonym. Es geht darum, verschiedene Textausschnitte zu lesen und diese zu ihrer Verständlichkeit und Inklusivität zu bewerten. Über untenstehenden Link geht es zur Google Form der Umfrage:}\\\\
\texttt{LINK TO SURVEY''}
\end{tabularx}
\vspace{0.3cm}
\end{tiny}

English translation: \\\\
\begin{tiny}
\vspace{0.3cm}
\begin{tabularx}{0.95\columnwidth}{X}
\texttt{``The Department of Computational Linguistics (University of Zurich) and Textshuttle work on text generation systems that output (more) gender-fair language and want to explore in a survey how different texts are perceived with respect to gender-fair language.}\\\\
\texttt{The survey takes 10-15 minutes and is anonymous. The goal is to read different text excerpts and to rate them according to their understandability and their inclusivity. Following the link below you can reach the Google Form for the survey:}\\\\
\texttt{LINK TO SURVEY''}
\end{tabularx}
\vspace{0.3cm}
\end{tiny}

\section{Intended Use}
\label{app:intended_use}

\begin{itemize}
    \item The models presented in this paper are intended to rewrite biased text with possible gender-fair forms; they are \textit{not} intended to identify a person's gender nor to prescribe a particular gender-fair form.
    \item The models are primarily trained for research purposes, showing that gender-fair rewriting models can be trained without language-specific handwritten rules.
    \item While our survey with potential beneficiaries highlights that gender-fair rewriting models -- even though not error-free -- may also be beneficial in real-world applications, we caution that they should be thoroughly tested by potential users before being deployed outside of research contexts.
\end{itemize}

\end{document}